\documentclass[11pt]{article}
\usepackage{coling2018}
\usepackage{times}
\usepackage{url}
\usepackage{latexsym}
\usepackage{amsmath}
\usepackage{multirow}
\usepackage[T1]{fontenc}
\usepackage{lmodern}
\usepackage{graphicx}

\usepackage{hyperref}

\title{Predicting and Analyzing Law-Making in Kenya}

\author{Oyinlola Babafemi \\
   Data Duality \\
   {\tt babsoyinlola@gmail.com} \\\And
   Adewale Akinfaderin \\
   Data Duality \\
   {\tt waleakinfaderin@gmail.com} \\}

\date{}

\begin{document}
\maketitle
\begin{abstract}
Modelling and analyzing parliamentary legislation, roll-call votes and order of proceedings in developed countries has received significant attention in recent years. In this paper, we focused on understanding the bills introduced in a developing democracy, the Kenyan bicameral parliament. We developed and trained machine learning models on a combination of features extracted from the bills to predict the outcome - if a bill will be enacted or not. We observed that the texts in a bill are not as relevant as the year and month the bill was introduced and the category the bill belongs to.
\end{abstract}
\blfootnote{4th Widening NLP Workshop, Annual Meeting of the Association for Computational Linguistics, ACL 2020}
\section{Introduction}

Policy development and law-making affect millions of people. It is important that there is transparency and openness in this decision making process. The rationale behind this work is to give insights to what happens in the Kenyan parliament and possible factors that might influence the verdict of bills. In Kenya bicameral parliament (the Senate and National Assembly), the legislative process goes through five phases: the proposed bill is published in the \emph{Kenya Gazette\footnote{Kenya Gazette is an official publication of the government of the Republic of Kenya}}, first reading, second reading, the appropriate committee meets to consider the amendment and finally, third and last reading~\cite{Goi:17}. After these phases, it is signed into law (or not) by the President of Kenya. Previous works have used word vectors and machine learning to estimate the probability that a United State congressional bill will survive the congressional committee and become law and, to predict policy changes in China~\cite{Nay:17,Tae:12,Cha:18}. Although data from debates and votes could not be obtained for this work, hand-crafted features and word vector representations of the texts in bills were used to predict if they will be enacted or not enacted.  

\section{Data and Methodology}

460 Kenyan national assembly and senate bills introduced between 2009 and 2019 were downloaded from the \emph{Kenya Gazette} website\footnote{http://kenyalaw.org/kl/index.php?id=9091} and the corresponding metadata scraped. Of these bills, 395 were not enacted while only 65 bills were passed into law. The highest number of bills introduced in a single year is 88 which was in 2012 and Aden Duale, the Majority Leader of the National Assembly of Kenya under the Jubilee Party introduced about 24\% of the bills retrieved.

\begin{figure}[!h]
  \centering
  \includegraphics[width=1\linewidth]{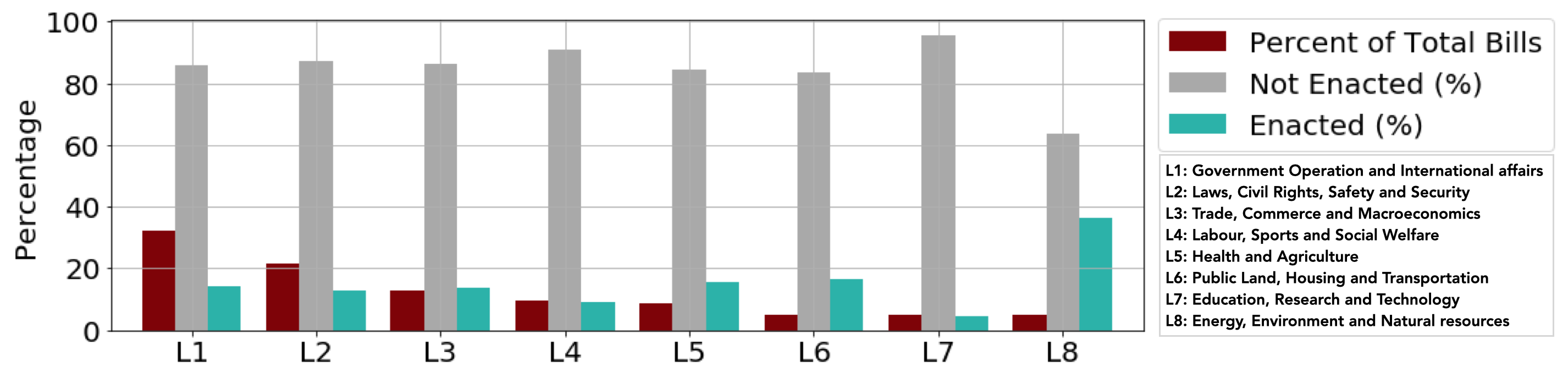}
  \caption{Percent distribution of bills and their corresponding enacted and not enacted percentages. L1 to L8 represent the bill labels. This is part of the features in our model.} 
    \label{figure:distribution}
\end{figure}

\subsection{Data Pre-processing and Feature Engineering}

To develop our model, we extracted information from the dataset and engineered new features. Some of the features we used are: the category of a bill - inspired by the socio-economic labels in~\cite{Aki:19}, election year - a binary feature that represents if a bill was introduced in an election year or not, sponsor 1 - binary feature for a bill sponsored by Aden Duale or others, sponsor 2 - binary feature for a bill sponsored by members of the parliament or, attorney generals and ministers, year and month a bill was introduced, the length of bill title, word vectors - word vectors for the bill titles and texts using a \emph{100-dimensional} GloVe~\cite{Pen:14} pre-trained word vectors\footnote{Word vectors trained on the bills did not perform as well as GloVe}, the difference between the current year and the year the bill was introduced. Figure \ref{figure:distribution} represents the distribution of the bills for each category label and the percentage of bills enacted and not enacted for each category. To solve the class imbalance problem caused by the ratio of bills enacted to not enacted, we oversampled the minority class in our training set (bills enacted) using SMOTE: Synthetic Minority Over-sampling Technique~\cite{Chw:02}.

\subsection{Model}

With a 70:30 train-test split on our data, we employed Logistic Regression and Support Vector Machine models to obtain baseline results before proceeding to stack these models as base learners and then used another Logistic Regression classifier as the meta-learner in a bid to obtain better results. %An initial baseline was obtained using Logistic Regression and non-textual features (without bills title and text) after which other textual features were introduced.%
Although the accuracy obtained from the results of all three models were very impressive, we focused on other metrics such as the F1-score, precision, recall, AUC (area under the curve) and brier score to analyse our results\footnote{https://github.com/BabafemiOyinlola/Predicting-Law-Making-in-Kenya}. The results are displayed in Table \ref{table:results}. While handling the data imbalance problem improved the base models with a 5\% and 11\% increase in the precision of both classes, the results obtained for the enacted class after oversampling remained the same. However, we obtained impressive results for predicting that a bill will not be enacted. For better context, this means when a new bill is proposed and fed to the final model with all relevant aforementioned features, the model is 81\% accurate in predicting if the bill will be passed into law or not. The precision and recall are 65\% and 71\% respectively.

\begin{table}[!h]
\centering
\begin{tabular}{l|c|c|c|c|c|c}
\hline
\textbf{Model} & \textbf{F1} & \textbf{Precision} & \textbf{Recall} & \textbf{AUC} & \textbf{Brier Loss} & \textbf{Accuracy} \\ \hline
Logistic Regression & 0.65 & 0.63 & 0.70 & 0.72 & 0.17 & 0.79 \\ 
Support Vector Machine & 0.63 & 0.62 & 0.69 & \textbf{0.74} & 0.17 & 0.77 \\
Stacked Ensemble & \textbf{0.67} & \textbf{0.65} & \textbf{0.71} & 0.73 &\textbf{0.15} & \textbf{0.81} \\ \hline
\end{tabular}
\caption{Evaluation metrics with results (average for both classes). For Brier score, lower is better.}
\label{table:results}
\end{table}

\begin{figure}[!h]
  \centering
  \includegraphics[width=1\linewidth]{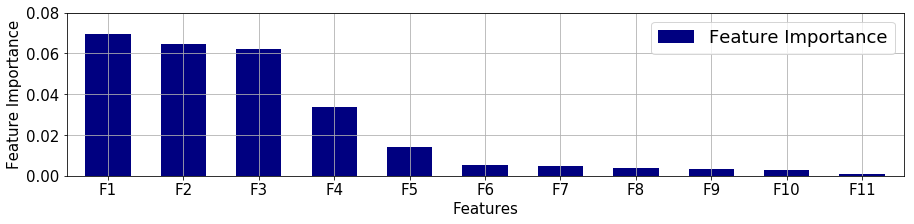}
  \caption{Feature importance plot displaying the relative importance of each feature in making a prediction. The features F1 to F11 represents the following: \textbf{F1}:\emph{Month}, \textbf{F2}:\emph{Label}, \textbf{F3}:\emph{Year Introduced}, \textbf{F4}:\emph{Text Length}, \textbf{F5}:\emph{Sponsor (Aden Duale or Not)}, \textbf{F6}:\emph{Sponsor2 (Executive or Legislator)}, \textbf{F7}:\emph{Title Length}, \textbf{F8}:\emph{Title Word Vector}, \textbf{F9}:\emph{Year Difference}, \textbf{F10}: \emph{Election Year or Not}, \textbf{F11}:\emph{Text Word Vectors}\}} 
    \label{figure:modelperformance}
\end{figure}

By inspecting the model to understand different features, we observed that the most important features that contribute to the final verdict were month, category and year introduced respectively (Figure \ref{figure:modelperformance}). Surprisingly, the title of a bill was a more important feature than the entire text, which is the least important feature - this raises a speculation that not all bills might be read thoroughly by members since the bills have similar titles which might cause voting parties to treat as bills previously introduced to the parliament. In addition, further experiments carried out using bag of words as an alternative representation for the textual features only confirmed that the text of a bill is not very pertinent to decision making. This suggests that there might be other factors considered in the parliament not accounted for here.

\section{Conclusion}

We presented simple baselines for predicting the chance that a new bill introduced will be enacted and, can adequately predict that a bill introduced into the Kenyan parliament will not be enacted (0.86 and 0.41 F1 scores for bills not enacted and bills enacted respectively). Avenues for improving this work include gathering more bills from earlier years, exploring other metadata like parliamentary debates and identifying dynamic structural factors in the behavioral patterns of the Kenya legislature.

% \section*{Acknowledgements}

% Anonymized

\section*{Acknowledgements}

 The authors are thankful to the anonymous reviewers for their valuable feedback, Ahmed Baruwa for data labelling and scraping and the Data Science Nigeria team for their support. This project is funded by the Artificial Intelligence 4 Development (AI4D) program as part of the 1st AI4D-Africa Innovation Call for Proposals. AI4D is sponsored by the Canadian International Development Research Centre (IDRC), the Knowledge for All Foundation (K4A) and the Swedish International Development Cooperation Agency (SIDA).

%\bibliography{coling2018}

\end{document}